\def\year{2018}\relax
\documentclass[letterpaper]{article} 
\usepackage{aaai18}  
\usepackage{times}  
\usepackage{helvet}  
\usepackage{courier}  
\usepackage{url}  
\usepackage{graphicx}  
\usepackage{amsmath}
\makeatletter
\def\hlinewd#1{%
  \noalign{\ifnum0=`}\fi\hrule \@height #1 \futurelet
   \reserved@a\@xhline}
\makeatother
\begin{document}
%

\title{Table-to-text Generation by Structure-aware Seq2seq Learning}

\author{Tianyu Liu, Kexiang Wang, Lei Sha, Baobao Chang \and Zhifang Sui \\
Key Laboratory of Computational Linguistics, Ministry of Education, \\School of Electronics Engineering and Computer Science, Peking University, Beijing, China\\
  {\tt \{tianyu0421, wkx, shalei, chbb, szf\}@pku.edu.cn}}
\maketitle
\begin{abstract}

Table-to-text generation aims to generate a description for a factual table which can be viewed as a set of field-value records.
To encode both the content and the structure of a table,
we propose a novel structure-aware seq2seq architecture 
which consists of field-gating encoder and description generator with dual attention. 
In the encoding phase, we update the cell memory of the LSTM unit by a field gate and its corresponding field value 
in order to incorporate field information into table representation.
In the decoding phase, dual attention mechanism which contains word level attention and field level attention is proposed 
to model the semantic relevance between the generated description and the table.
We conduct experiments on the \texttt{WIKIBIO} dataset which contains over 700k biographies and corresponding infoboxes from Wikipedia. 
The attention visualizations and case studies show that our model is capable of generating coherent and informative descriptions based on the comprehensive understanding of both the content and the structure of a table.
Automatic evaluations also show our model outperforms the baselines by a great margin. Code for this work is available on 
https://github.com/tyliupku/wiki2bio.
\end{abstract}

\section{Introduction}
\begin{figure}[t]
\centering
\includegraphics[width=0.6\linewidth]{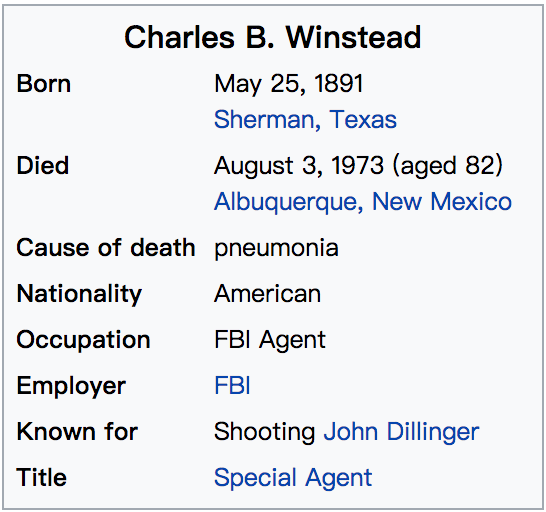}
\begin{tabular}{@{}c@{}@{}c@{}@{}c@{}@{}c@{}}
\end{tabular}
\caption{The Wikipedia infobox of Charles Winstead, the corresponding introduction on his wiki page reads ``Charles Winstead (1891 {}- 1973) was an FBI agent in the 1930s -{} 40s, famous for being one of the agents who shot and killed John Dillinger.''.}\label{table1}
\end{figure}

Generating natural language description for a structured table is an important task for text generation from structured data.
Previous researches include weather forecast based on a set of  weather records  \cite{liang2009learning} and sportscasting based on temporally ordered events  \cite{chen2008learning}. 
However, previous work models the structured data in the limited pre-defined schemas. For example, a weather record \textit{rainChance(time:06:00-21:00, mode:SSE, value:20)} is represented by a fixed-length one-hot vector by its record type, record time, record value and record value.
To this end, we focus on table-to-text generation which involves comprehensive representation for the complex structure of a table rather than pre-defined schemas.
In contrast to previous work experimented on small datasets which contain only a few tens of thousands of records such as 
\texttt{WEATHERGOV} \cite{liang2009learning} and \texttt{ROBOCUP} \cite{chen2008learning}, 
we focus on a more challenging task to generate biographies based on the Wikipedia infoboxes. 
As shown in Fig \ref{table1}, a biographic infobox is a fixed-format table that describes a person with many field-value records like (Name,\textit{ Charles B. Winstead}), (Nationality, \textit{American}), (Occupation, \textit{FBI Agent}), etc. 
We utilize \texttt{WIKIBIO} dataset proposed by \citeauthor{lebret2016neural} \shortcite{lebret2016neural} which contains 700k biographies from Wikipedia, with 400k words in total as the benchmark dataset. 

Previous work has made significant progress on this task.
\citeauthor{lebret2016neural} \shortcite{lebret2016neural} proposed a statistical n-gram model with local and global conditioning on a Wikipedia infobox. 
However the field content of a record is likely to be a sequence of words, the statistical language model is not good at capturing long-range dependencies between words.  
\citeauthor{mei2015talk} \shortcite{mei2015talk}
proposed a selective generation method based on an encoder-aligner-decoder framework. 
The model utilizes a sparse one-hot vector to represent a weather record.
However it's inefficient to represent the complex structure of a table by one-hot vectors.

We propose a structure-aware sequence to sequence (seq2seq) generation framework to model both content and structure of the table by local and global addressing. 
When a human writes a biography for a person based on the related Wikipedia infobox, he will firstly determine which records in the table should be included in the introduction and how to arrange the order of these records before wording. 
After that, the writer will further consider which words or phrases in the table should be more focused on to paraphrase.
We summarize the two phases of generation as two scopes of addressing: local and global addressing.
\textbf{Local addressing} determines which particular word in the table should be focused on while generating a piece of description at certain time step. 
However, the word level addressing can not fully address the table-to-text generation problem as the factual tables usually have complex structures which might confuse the generator.
\textbf{Global addressing} is proposed to determine which records of the table should be more focused on while generating corresponding description.
Global addressing is necessary as the description of a table may not cover all the records. For example, the `cause of death' field in Fig \ref{table1} is not mentioned in the description.
Furthermore, the order of records in the tables may not always be homogeneous. For example, we can introduce a person as an order of his (Birth-Death-Nationality-Occupation) according to his Wikipedia infobox. However the other infoboxes may be arranged as (Occupation-Nationality-Birth-Death).
Local addressing is realized by content encoding of the LSTM encoder and word level attention 
while global addressing is realized by field encoding of the field-gating LSTM variation and field level attention in our model.

The structure-aware seq2seq architecture we proposed exploits encoder-decoder framework using long short-term memory (LSTM) \cite{hochreiter1997long} units with local and global addressing on the structured table.   
In the encoding phase, our model first encodes the sets of field-value records in the table by integrating field information and content representation.
To make better use of field information, we add a field gate to the cell state of the encoder LSTM unit to incorporate the field embedding into the structural representation of the table.
The model next employs a LSTM decoder to generate natural language description by the structural representation of the table.
In the decoding phase, we also propose a novel dual attention mechanism which consists of two parts: word-level attention for local addressing and field-level attention for global addressing.

Our contributions are three-fold:
(1) We propose an end-to-end structure-aware encoder-decoder architecture to encode field information into the representation of a structured table.
(2) Field-gating encoder and dual attention mechanism are proposed to operate local and global addressing between the content and the field information of a structured table. 
(3) Experiments on \texttt{WIKIBIO} dataset show that our model achieves substantial improvement over baselines.
 

\section{Related Work}
Most generation systems can be divided into two independent modules: 
(1)\textit{content selection} involves choosing a subset of relevant records in a table to talk about.
(2)\textit{surface realization} is concerned with generating natural language descriptions for this subset.

Many approaches have been proposed to learn the individual modules. 
For content selection module, one approach builds a content selection model by aligning records and sentences \cite{barzilay2005collective,duboue2002content}. 
A hierarchical semi-Markov method is proposed by \cite{liang2009learning} which first associates the text sequences to corresponding records and then generates corresponding descriptions from these records.
Surface realization is often treated as a concept-to-text generation task from a given representation. 
\citeauthor{reiter2000building} \shortcite{reiter2000building}, \citeauthor{walker2001spot} \shortcite{walker2001spot} and \citeauthor{stent2004trainable} \shortcite{walker2001spot} utilize various linguistic features to train sentence planners for sentence generation. 
Context-free grammars are also used to generate natural language sentences from formal meaning representations \cite{lu2011probabilistic,belz2008automatic}. 
Other effective approaches include hybrid alignment tree \cite{kim2010generative}, tree conditional random fields \cite{lu2009natural}, tree adjoining grammar \cite{gyawali2016surface} and template extraction in a log-linear framework \cite{angeli2010simple}.
Recent work combines content selection and surface realization in a unified framework \cite{ratnaparkhi2002trainable,konstas2012unsupervised,konstas2013global,DBLP:journals/corr/abs-1709-00155}

Our model borrowed the idea of representing a structured table by its field and content information from \cite{lebret2016neural}. However, 
their n-gram model is inefficient to model long-range dependencies while generating descriptions.
\citeauthor{mei2015talk} \shortcite{mei2015talk} also proposed a seq2seq model with an aligner between weather records and weather broadcast. The model used one-hot encoding to represent the weather records as they are relatively simple and highly structured. However, the model is not capable to represent the tables with complex structure like Wikipedia infoboxes.    
\section{Task Definition}


We model the table-to-text generation in an end-to-end structure-aware seq2seq framework. The given table $T$ can be viewed as a combination of $n$ field-value records \{ $R_1, R_2, \cdots, R_n$\}. Each record $R_i$ consists of a sequence of words \{ $d_1, d_2, \cdots, d_m$\} and their corresponding field represent \{ $Z_{d_1}, Z_{d_2}, \cdots, Z_{d_m}$\}.

The output of the model is the generated description $S$ for table $T$ which contains $p$ tokens \{$w_1, w_2, \cdots, w_p $\} with $w_t$ being the word at time $t$. We formulate the table-to-text generation as the inference over a probabilistic model.
The goal of the inference is to generate a sequence $w_{1:p}^*$ which maximizes $P(w_{1:p}|R_{1:n})$.  

\begin{equation}
	w_{1:p}^* = \arg \max \limits_{w_{1:p}} \prod_{t=1}^{p} P(w_t|w_{0:t-1},R_{1:n})
	\label{io1}
\end{equation}

\begin{figure}[]
\centering
\includegraphics[width=0.9\linewidth]{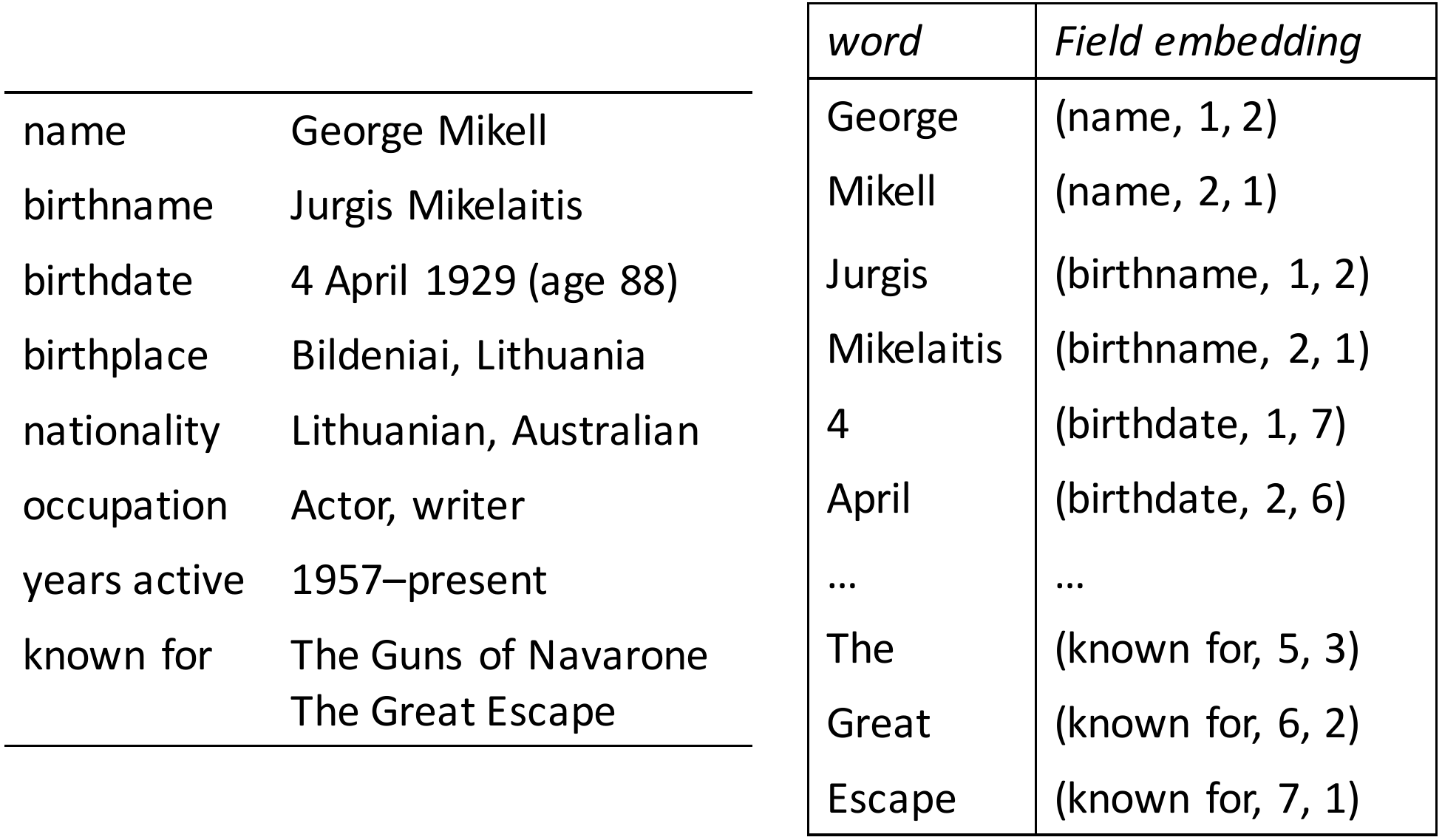}
\caption{The wiki infobox of \textit{George Mikell} (left) and the table of its field representation (right).}\label{table2}
\end{figure} 

\section{Structure-aware Seq2seq}

\subsection{Field representation}
A Wikipedia infobox can be viewed as a set of field-value records, in which values are sequences or segments of words corresponding to certain fields.
The structural representation of an infobox consists of context embedding and field embedding. 
The context embedding is formulated as an embedding for a segment of words in the field content. 
The field embedding is a key point to label each word in the field content by its corresponding field name and occurrence in the table.  
\citeauthor{lebret2016neural} \shortcite{lebret2016neural} represented the field embeddding $Z_w=\{ f_w; p_w\}$ for a word $w$ in the table with corresponding field name $f_w$ and position information $p_w$. 
The position information can be further represented as a tuple ($p_w^+$, $p_w^-$) which includes the positions of the token $w$ counted from the begining and the end of the field respectively. So the field embedding of token $w$ is extended to a triple:
\begin{equation}
  Z_w= \{ f_w; p_w^+; p_w^-\}
\end{equation}
As shown in Fig \ref{table2}, the infobox of \textit{George Mikell} contains several field-value records, the field content for the record (birthname, \textit{Jurgis Mikelatitis}) is `\textit{Jurgis Mikelatitis}'. The word `\textit{Jurgis}' is the first token counted from the beginning of the field `birthname'  and also the second token counted from the end. 
So the field embedding for the word `Jurgis' is described as $\{ birthname; 1; 2\}$.
Each token in the table has an unique field embedding even if there exists two same words in the same field due to the unique (field, position) pair. 
\subsection{Field-gating Table Encoder}
The table encoder aims to encode each word $d_j$ in the table together with its field embedding $Z_{d_j}$ 
into the hidden state $h_j$ using LSTM encoder. 
We present a novel field-gating LSTM unit
to incorporate field information into table encoding.
LSTM is a recurrent neural network (RNN) architecture which uses a vector of cell state $c_t$ and a set of element-wise multiplication gates to control how information is stored, forgotten and exploited inside the network. Following the design for an LSTM cell in \cite{graves2013speech} , the architecture used in the table encoder is defined by following equations: 
\begin{equation}
\begin{pmatrix}
i_t\\
f_t\\
o_t\\
\hat{c}_t
\end{pmatrix}
= 
\begin{pmatrix}
sigmoid\\
sigmoid\\
sigmoid\\
tanh
\end{pmatrix}
W_{4n,2n}^c
\begin{pmatrix}
d_t\\
h_{t-1}
\end{pmatrix}
\label{encoder_eq1}
\end{equation}

\begin{equation}
	c_t = f_t \odot c_{t-1} + i_t \odot \hat{c}_t
\label{encoder_eq2}
\end{equation}
\begin{equation}
	h_t = o_t \odot tanh(c_t)
\label{encoder_eq3}
\end{equation}
where $i_t, f_t, o_t \in [0,1]^n$ are input, forget and output gates respectively, and $\hat{c}_t$ and $c_t$ are proposed cell value and true cell value in time $t$. $n$ is the hidden size. 

To make better understanding of the structure of a table, the field information should also be encoded into the encoder. 
One simple way is to take the concatenation of word embedding and corresponding field embedding as the input for the vanilla LSTM unit. Actually, the method is indeed proved to be useful in our experiments and serves as a baseline for comparison.  
However, the concatenation of word embedding and field embedding only treats the field information as an additional label of certain token which loses the structural information of the table. 

To better encode the structural information of a table, we propose a field-gating variation on the vanilla LSTM unit to update the cell memory by a field gate and its corresponding field value. The field-gating cell state is described as follows:   
 \begin{equation}
\begin{pmatrix}
l_t\\
\hat{z}_t
\end{pmatrix}
= 
\begin{pmatrix}
sigmoid\\
tanh
\end{pmatrix}
W_{2n,2n}^d
\begin{pmatrix}
z_t
\end{pmatrix}
\end{equation}

\begin{equation}
	c_t^{'} = f_t \odot c_{t-1} + i_t \odot \hat{c}_t + l_t \odot \hat{z}_t
\end{equation}
where $z_t$ is the field embedding described before, $l_t\in [0,1]^n$ is the field gate to determine how much field information should be kept in the cell memory, $\hat{z}_t$ is the proposed field value corresponding to field gate. The cell state $c_t^{'}$ is updated from the original $c_t$ by incorporating field information of the table.

\begin{figure*}[]
\centering
\includegraphics[width=0.9\linewidth]{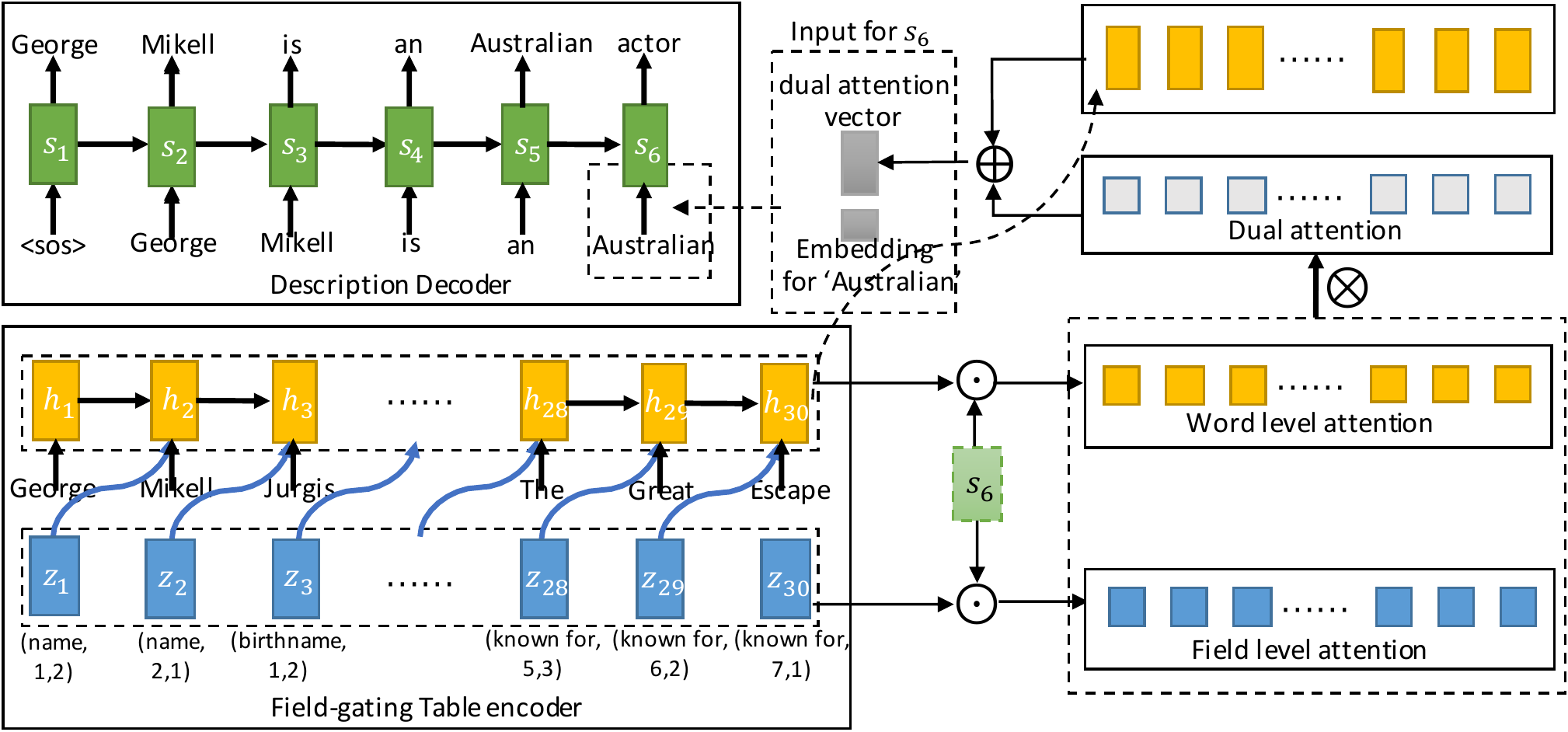}
\caption{The overall diagram of structure-aware seq2seq architecture for generating description for \textit{George Mikell} in Fig \ref{table2}.}\label{mmodel}
\end{figure*}

\subsection{Description Decoder with Dual Attention}
To conduct local and global addressing towards the structured table, we use LSTM architecture with dual attention mechanism as our description generator. As defined in the equation \ref{io1}, the generated token $w_t$  at time $t$ in the decoder is predicated based on all the previously generated tokens $w_{<t}$ before $w_t$, the hidden states $H=\left\{h_t\right\}_{t=1}^{L}$ of the table encoder and the field embeddings $Z=\left\{z_t\right\}_{t=1}^{L}$. To be more specific:
\begin{equation}
P(w_t|H,Z,w_{<t}) = softmax(W_s\odot g_t)
\end{equation}
\begin{equation}
g_t = tanh(W_t[s_t,a_t])
\end{equation}
\begin{equation}
s_t = LSTM(w_{t-1}, s_{t-1})
\end{equation}

where $s_t$ is the $t$-th hidden state of the decoder calculated by the LSTM unit. The computational details can be referred in Equation \ref{encoder_eq1}, \ref{encoder_eq2} and \ref{encoder_eq3}. $a_t$ is the attention vector which is widely used in many applications \cite{xu2015show,luong2014addressing,DBLP:conf/acl/MaSXWLS17}. 
Vanilla attention mechanism is proposed to encode the semantic relevance between the encoder states $\left\{h_t\right\}_{t=1}^{L}$ and and the decoder state $s_t$ at time $t$. The attention vector is usually represented by the weighted sum of encoder hidden states.
\begin{equation}
	  \alpha_{t_i}=\frac{e^{g(s_{t},h_{i})}}{\sum_{j=1}^{N}{e^{g(s_{t},h_{j})}}}; a_t = \sum_{i=1}^L \alpha_{t_i} h_i
\end{equation}
where $g(s_{t},h_{i})$ is a relevant score between decoder hidden state $s_t$ and encoder hidden state $h_i$. There are many different ways to calculate the relevant scores. In our paper, we use the following dot product to measure the similarity between $s_t$ and $h_i$. $W_s,W_t,W_p,W_q$ are all model parameters. 
\begin{equation}
	g(s_{t},h_{i}) = tanh(W_ph_i) \odot tanh(W_qs_t)
\label{de01}
\end{equation}

However, the word level attention described above can only capture the semantic relevance between generated tokens and the content information in the table, ignoring the structure information of the table. To fully utilize the structure information, we propose a higher level attention over generated tokens and the field embedding of the table. 
Field level attention can locate the particular field-value record which
should be focused on while generating next token in the description  by modeling the relevance between all field embeddings $\left\{z_t\right\}_{t=1}^{L}$ and the decoder state $s_t$ at $t$-th time. 
Field level attention weight $\beta_{t_i}$ is presented as Equation \ref{ep33}.
We use the same relevant score function $g(s_{t},z_{i})$ as equation \ref{de01}. 
Dual attention weight $\gamma_t$ is the element-wise production between field level attention weight $\beta_t$ and word level attention weight $\alpha_t$. The dual attention vector $a_t^{'}$ is updated as the weighted sum of encoder states $\left\{h_t\right\}_{t=1}$ by $\gamma_t$ (Equation \ref{dualeq}):
\begin{equation}
\beta_{t_i}=\frac{e^{g(s_{t},z_{i})}}{\sum_{j=1}^{N}{e^{g(s_{t},z_{j})}}}
\label{ep33}
\end{equation}
\begin{equation}
	g(s_{t},z_{i}) = tanh(W_xz_i) \odot tanh(W_ys_t)
\end{equation}
\begin{equation}
	\gamma_{t_i} =  \frac{\alpha_{t_i} \cdot \beta_{t_i}}
	{\sum_{j=1}^{N}{\alpha_{t_j} \cdot \beta_{t_j}}} ;
	a_t^{'} = \sum_{i=1}^L \gamma_{t_i} h_i
\label{dualeq}
\end{equation}

Furthermore, we utilize a post-process operation for the generated unknown (UNK) tokens to alleviate the out-of-vocabulary (OOV) problem. We replace a specific generated UNK token with the most relevant token in the corresponding table according to the related dual attention matrix. 

\subsection{Local and Global Addressing}
Local and global addressing determine which part of the table should be more focused on in different steps of description generation. The two scopes of addressings play a very important role in understanding and representing the inner-structure of a table. 
Next we will introduce how our model conducts local and global addressing on table-to-text generation with the help of Fig \ref{mmodel}.

\begin{table*}[htbp]
\small
\centering
\begin{tabular}{ccccccc}
  \hline 
  & \# tokens per sentence & \# table token per sent.  & \# tokens per table & \# fields per table \\
Mean & 26.1 & 9.5 & 53.1 & 19.7 \\\hline
\end{tabular}
\caption{\label{corpus-stat}Statistics of \texttt{WIKIBIO} dataset.}
\end{table*}

\begin{table*}[htbp]
\small
\centering
\begin{tabular}{ccccccc}
  \hline 
Word dimension & Field dimension & Position dimension & Hidden size & Batch size & Learning rate & Optimizer\\
400 & 50 & 5 & 500 & 32 & 0.0005 & Adam \\\hline
\end{tabular}
\caption{\label{parameter}Parameter settings of our experiments.}
\end{table*}

\textbf{Local addressing}: A table can be treated as a set of field-value records. Local addressing tends to encode the table content inside each record. 
The value in each field-value record is a sequence of words
which contains 2.7 tokens on average.
Some records in the Wikipedia infoboxes even contain several phrases or sentences. 
Previous models which used one-hot encoding or statistical language model to encode field content are inefficient to capture the semantic relevance between words inside a field. 
The seq2seq structure itself has a strong ability to model the context of a piece of words. For one thing, the LSTM encoder can capture long-range dependencies between words in the table. For another, the word level attention of the proposed dual attention mechanism can also build a connection between the words in the description and the tokens in the table. The generated word `\textit{actor}' in Fig \ref{mmodel} refers to the word `\textit{actor}' in the `Occupation' field.

\textbf{Global addressing}: 
The goal of local addressing is to represent inner-record information while global addressing aims to model inter-record relevance within the table.   
For example, it's noteworthy that the generated token `\textit{actor}' in Fig \ref{mmodel} is mapped to the `occupation' field in Table \ref{table2}.
  
Field-gating table representation and field level attention mechanism are proposed for global addressing.
For table representation, we encode the structure of a table by incorporating field and position embedding into table representation apart from word embedding.
The position of a token in the field content of a table is determined only by its field and position information. Even two same words in the table can be distinguished by their field and position. 
We propose a novel field-gating LSTM to incorporate the field embedding into the cell memory of LSTM unit.

Furthermore, the information in a table is likely to be redundant. Some records in the table are unimportant or even useless for generating description. We should make appropriate choices on selecting useful information from all the table records.
The order of records may also influence the performance of generation \cite{vinyals2015order}. 
We should make it clear which records the token to be generated is focused on by global addressing between the field information of a table and its description. 
The field level attention of dual attention mechanism is introduced to determine which field the generator focused on in certain time step. 
Experiments show that our dual attention mechanism is of great help to generate description from certain table and insensible to different orders of table records.

\section{Experiments} 
We first introduce the dataset, evaluation metrics and experimental setups in our experiments. Then we compare our model with several baselines. After that, we assess the performance of our model on table-to-text generation. Furthermore, we also conduct experiments on the disordered tables to show the efficiency of global addressing mechanism.     
\subsection{Dataset and Evaluation Metrics}
We use  \texttt{WIKBIO} dataset proposed by \citeauthor{lebret2016neural} \shortcite{lebret2016neural} as the benchmark dataset. \texttt{WIKBIO} contains 728,321 articles from English Wikipedia (Sep 2015). The dataset uses the first sentence of each article as the description of the corresponding infobox. Table \ref{corpus-stat} summarizes the dataset statistics: on average, the tokens in the table (53.1) are twice as long as those in the first sentence (26.1). 9.5 tokens in the description text also occur in the table. The corpus has been divided in to training (80\%), testing (10\%) and validation (10\%) sets. 

We assess the generation quality automatically with BLEU-4 and ROUGE-4 (F measure)\footnote{We use standard scripts NIST mteval-v13a.pl (for BLEU), and rouge-1.5.5 (for ROUGE).} .

\subsection{Baselines}
We compare the proposed structure-aware seq2seq model with several statistical language models and the vanilla encoder-decoder model.
The baselines are listed as follows:
\begin{itemize}
	\item \textbf{KN}: The Kneser-Ney (KN) model is a widely used language model proposed by \citeauthor{heafield2013scalable} \shortcite{heafield2013scalable}. We use the KenLM toolkit to train 5-gram models without pruning.
	\item \textbf{Template KN}: Template KN is a KN model over templates which also serves as a baseline in \cite{lebret2016neural}. The model replaces the words occurring in both the table and the training sentences with a special token reflecting its field. The introduction section of the table in Fig \ref{table2} looks as follows under this scheme: `` \texttt{name\_1 name\_2} (born \texttt{birthname\_1 
... birthdate\_3}) is a Lithuanian-Australian \texttt{occupation\_1} and \texttt{occupation\_3} best known for his performances in \texttt{known\_for\_1 ... known\_for\_4} (1961) and \texttt{known\_for\_5 ... known\_for\_7} (1963) ''. During inference, the decoder is constrained to emit words from the regular vocabulary or special tokens occurring in the input table. 
	\item  \textbf{NLM}: A naive statistical language model proposed by \cite{lebret2016neural} for comparison. The model uses only the field content as input without field and position information.
	\item \textbf{Table NLM}: The most competitive statistical language model proposed by \cite{lebret2016neural}, which includes local and global conditioning over the table by integrating related field and position embedding into the table representation.
	\item \textbf{Vanilla Seq2seq}: The vanilla seq2seq neural architecture is also provided as a strong baseline which uses the concatenation of word embedding, field embedding and position embedding as the model input. The model can operate local addressing over the table by the natural advantages of LSTM units and word level attention mechanism. 
\end{itemize}

\begin{figure*}[ht]
\centering
\includegraphics[width=0.95\linewidth]{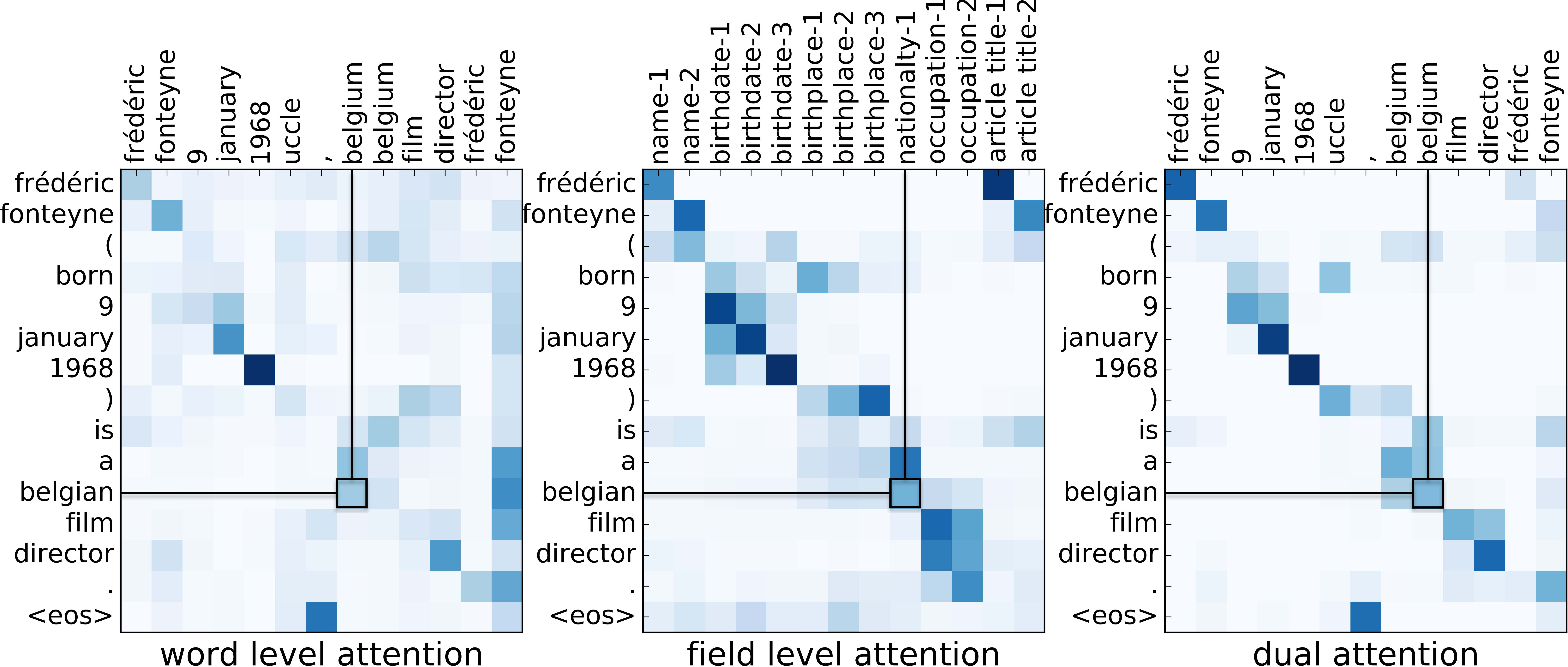}
\caption{An example of word level, field level and aggregated dual attention on generating the biography of Fr\'ed\'eric Fonteyne. Note there are two adjacent `belgium's in `birthplace-3' and  `nationality-1' field, respectively. The word level attention focuses improperly on the first `belgium' while generating `a belgian film director'. In contrast, the field level attention and dual attention can locate the second `belgium' properly by word-field modeling (marked in the black boxes).}\label{attshow}
\end{figure*}

\subsection{Experiment Setup}
In the table encoding phase, we use a sequence of word embeddings and their corresponding field and position embedding as input.
We select the most frequent 20,000 words in the training set as the word vocabulary. 
For field embedding, we select 1480 fields occurring more than 100 times from the training set as field vocabulary. Additionally, we filter all empty fields whose values are \textit{$\langle none \rangle$} while feeding field information to the network. 
We also limit the largest position number as 30. Any position number over 30 will be counted as 30.   

While generating description for the table, a special start token \textit{$\langle sos \rangle$} is feed into the generator in the beginning of the decoding phase. Then we use the last generated token as the input at the next time step.
A special end token \textit{$\langle eos \rangle$} is used to mark the end of decoding. We also restrict the generated text by a pre-defined max length to avoid redundant or irrelevant generation.
We also try \textbf{beam search} with beam size 2-10 to enhance the performance. 
We use grid search to determine the parameters of our model. The detail of model parameters is listed in Table \ref{parameter}.

\subsection{Generation Assessment}

\begin{table}[t]
	\begin{center}
		\begin{tabular}{lcc}		
		\hlinewd{1pt} \textbf{Model} & \textbf{BLEU} & \textbf{ROUGE} \\ \hline
		KN & 2.21 & 0.38 \\ 
		Template KN & 19.80 & 10.70 \\
		NLM & 4.17 $\pm$ 0.54 & 1.48 $\pm$ 0.23 \\ 
		Table NLM & 34.70 $\pm$ 0.36 & 25.80 $\pm$ 0.36  \\ \hline
		Seq2seq   & 42.06 $\pm$ 0.32 & 38.06 $\pm$ 0.36 \\
		+ field (concate) & 43.34 $\pm$ 0.37 & 39.84 $\pm$ 0.32\\
		+ pos (concate) & 43.65 $\pm$ 0.44 & 40.32 $\pm$ 0.23 \\ \hline
		Field-gating Seq2seq & 43.74 $\pm$ 0.23 & 40.53 $\pm$ 0.31 \\
		+ dual attention & \textbf{44.89 $\pm$ 0.33} & 41.21 $\pm$ 0.25 \\
		+ beam search (k=5) & 44.71& \textbf{41.65} \\
			\hlinewd{1pt} \\
		\end{tabular}
		\caption{BLEU-4 and ROUGE-4 for structure-aware seq2seq model (last three rows), statistical language model (first four rows) and vanilla seq2seq model with field and position input (three rows in the middle).}
		\label{res}
	\end{center}
\end{table}

The assessment for description generation is listed in Table \ref{res}. We have following observations:
(1) Neural network models perform much better than statistical language models. Even vanilla seq2seq architecture with word level attention outperform the most competitive statistical model by a great margin.
(2)The proposed structure-aware seq2seq architecture can further improve the table-to-text generation compared with the competitive vanilla seq2seq. Dual attention mechanism is able to boost the model performance by over 1 BLEU compared to vanilla attention mechanism.

\begin{figure*}[t]
\centering
\includegraphics[width=0.95\linewidth]{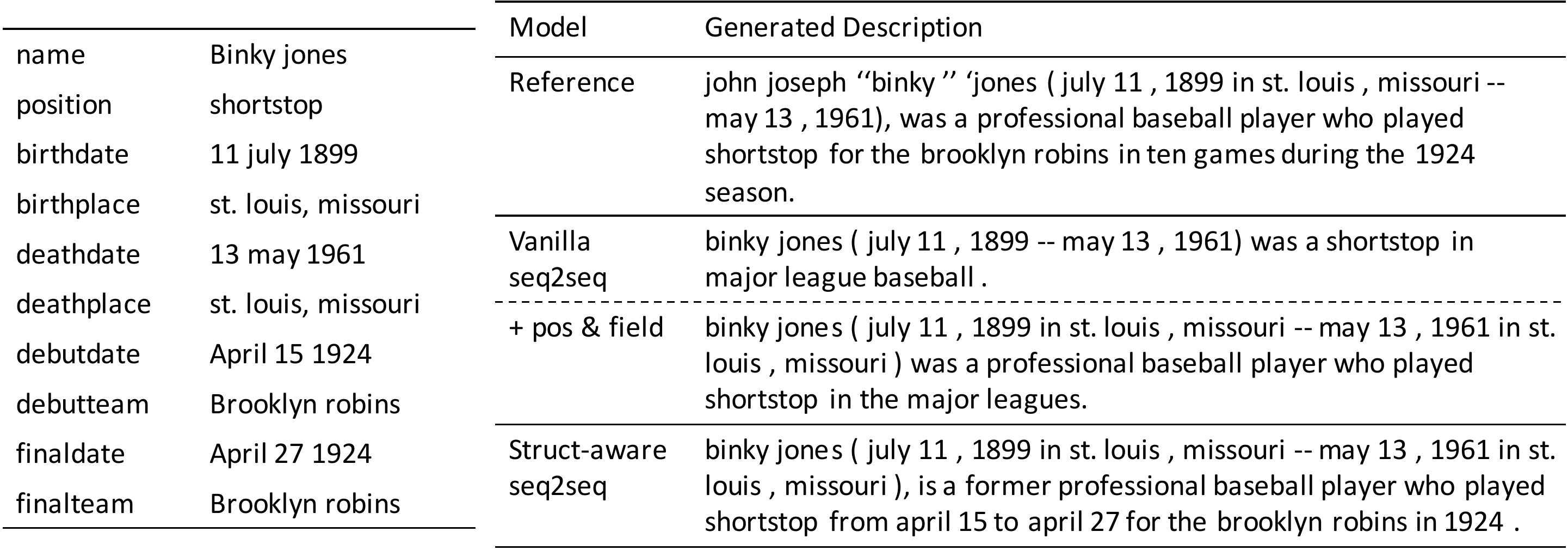}
\caption{The generated descriptions  for \textit{Binky Jones} and the corresponding reference in the Wikipedia. Our proposed struct-aware seq2seq model can generate more informative and accurate description compared to vanilla seq2seq model.}\label{case-study}
\end{figure*}

\subsection{Research on Disordered Tables}
We view a structured table as a set of field-value records and then feed the records into the generator sequentially as the order they are presented in the table.
The order of records can guide the description generator to produce an introduction in the pre-defined schemas \cite{vinyals2015order}.
However, not all the tables are arranged in the proper order. So global addressing between the generated descriptions and the records of the table is necessary for table-to-text generation.

Furthermore, the schemas of various types of tables differ greatly from each other. A biography about a politician may emphasize his or her social activities and working experience while a biography of a soccer player is likely to highlight which team he or she used to serve in or the performance in his or her career.
To cope with various schemas of different tables, it's essential to model inter-record information within the tables by global addressing.   
 
For these reasons, we propose a pair of disordered training and testing set based on \texttt{WIKIBIO} by randomly shuffling the records of a infobox. For example, the order of several records in a specific infobox is `name-birthdate-occupation-spouse', we randomly shuffle the table records as `occupation-name-spouse-birthdate', without changing the field content inside the `occupation', `name', `spouse' and `birthdate' records.

Table \ref{disorder} shows that 
all three neural network models perform not as good as before, which means the order of table records is an essential aspect for table-to-text generation.
However, the BLEU and ROUGE decreases on the structure-aware seq2seq model are much smaller than the other two models, which proves the efficiency of global addressing mechanism. 
\begin{table}[h]
	\begin{center}
		\begin{tabular}{lcc}		
		\hlinewd{1pt} \textbf{Model} & \textbf{BLEU} & \textbf{ROUGE} \\ \hline
		Seq2seq  & 40.04 (-2.02)& 36.85 (-1.21)\\
		+ field \& pos & 42.10 (-1.55) & 38.97 (-1.35)\\ \hline
		Structure-aware & \textbf{44.28 (-0.61)} & \textbf{40.79 (-0.42)} \\ 
			\hlinewd{1pt} \\
		\end{tabular}
		\caption{Experiments on the disordered tables to show the efficiency of global addressing.}
		\label{disorder}
	\end{center}
\end{table}

\section{Qualitative Analysis}
\subsection{Analysis on Dual Attention}
Dual attention mechanism models the relationship between the generated tokens and table content inside each record by word level attention while encoding the relevance of generated description and inter-record information within the table by field level attention.
The aggregation of word level attention and field level attention can  model more precise connection between the table and its generated description. 

Fig \ref{attshow} shows an example of the three attention mechanisms while generating a piece of description for \textit{Fr\'ed\'eric Fonteyne} based on his Wikipedia infobox.
We can find out that the name, birthdate, nationality and occupation information contained in the generated sentence can properly refer to the related table content by the aggregated dual attention.

\subsection{Case Study}
Fig \ref{case-study} shows the generated descriptions for different variants of our model based on the related Wikipedia infobox. 
All three neural network generators can produce coherent and understandable sentences with the help of local addressing mechanism. All of them contain the word `baseball' which is not directly mentioned in the infobox. 
It means the generators deduce from table content that \textit{Binky Jones} is a baseball player.

However, the two vanilla seq2seq models also generate `major league baseball' or `major leagues' which are not mentioned in the table and probably not correct. Vanilla seq2seq model without global addressing on the table just generates the most possible league in Wikipedia for a baseball player to play in. 

Furthermore, the two biographies generated by vanilla seq2seq model fail to contain the information from the infobox which team he served in, as well as the time period of his playing in that team.
The biography generated by our proposed structure-aware seq2seq model is able to cover nearly all the information mentioned in the table. The generated segment `who played shortstop from april 15 to april 27 for the brooklyn robins in 1924' (15 words) includes information in five fields of the table: `position', `debutdate', `finaldate', `debutteam' and `finalteam', which is achieved by the global addressing between the fields and the generated tokens.

\section{Conclusions}
We propose a structure-aware seq2seq architecture to encode both the content and the structure of a table for table-to-text generation.
The model consists of field-gating encoder and description generator with dual attention. We add a field gate to the encoder LSTM unit to incorporate the field information. 
Furthermore, dual attention mechanism which contains word level attention and field level attention can operate local and global addressing to the content and the structure of a table.
A series of visualizations, case studies and generation assessments show that our model outperforms the competitive baselines by a large margin.

\section{Acknowledgments}

Our work is supported by the National Key Research and Development Program of China under Grant No.2017YFB1002101 and project 61772040 supported by NSFC. The corresponding authors of this paper are Baobao Chang and Zhifang Sui.

\bibliography{paper}
\bibliographystyle{aaai}
\end{document}